\documentclass[conference]{IEEEtran}
\IEEEoverridecommandlockouts
\usepackage{cite}
\usepackage{stix}
\usepackage{amsmath,amssymb,amsfonts}
\usepackage{algorithmic}
\usepackage{graphicx}
\usepackage{textcomp}
\usepackage{xcolor}
\def\BibTeX{{\rm B\kern-.05em{\sc i\kern-.025em b}\kern-.08em
    T\kern-.1667em\lower.7ex\hbox{E}\kern-.125emX}}

\makeatletter
\newcommand{\linebreakand}{
\end{@IEEEauthorhalign}
\hfill\mbox{}\par
\mbox{}\hfill\begin{@IEEEauthorhalign}
}
\makeatother

\begin{document}

\title{A Closed-Loop Personalized Learning Agent Integrating Neural Cognitive Diagnosis, Bounded-Ability Adaptive Testing, and LLM-Driven Feedback\\
\thanks{Corresponding author: Chunyan Zeng, Email: cyzeng@hbut.edu.cn.}
}

\author{\IEEEauthorblockN{Zhifeng Wang}
	\IEEEauthorblockA{\textit{Faculty of Artificial Intelligence in Education} \\
		\textit{Central China Normal University}\\
		Wuhan 430079, China \\
		zfwang@ccnu.edu.cn}
	\and
	\IEEEauthorblockN{Xinyue Zheng}
	\IEEEauthorblockA{\textit{Faculty of Artificial Intelligence in Education} \\
		\textit{Central China Normal University}\\
		Wuhan, China \\
		zxy0624@foxmail.com}
	\linebreakand
	\IEEEauthorblockN{Chunyan Zeng}
	\IEEEauthorblockA{\textit{School of Electrical and Electronic Engineering} \\
		\textit{Hubei University of Technology}\\
		Wuhan 430068, China \\
		cyzeng@hbut.edu.cn}
}

\maketitle

\IEEEpubidadjcol

\begin{abstract}
As information technology advances, education is moving from one-size-fits-all instruction toward personalized learning. However, most methods handle modeling, item selection, and feedback in isolation rather than as a closed loop. This leads to coarse or opaque student models, assumption-bound adaptivity that ignores diagnostic posteriors, and generic, non-actionable feedback. 
To address these limitations, this paper presents an end-to-end personalized learning agent, EduLoop-Agent, which integrates a Neural Cognitive Diagnosis model (NCD), a Bounded-Ability Estimation Computerized Adaptive Testing strategy (BECAT), and large language models (LLMs). The NCD module provides fine-grained estimates of students' mastery at the knowledge-point level; BECAT dynamically selects subsequent items to maximize relevance and learning efficiency; and LLMs convert diagnostic signals into structured, actionable feedback. Together, these components form a closed-loop framework of ``Diagnosis--Recommendation--Feedback.''
Experiments on the ASSISTments dataset show that the NCD module achieves strong performance on response prediction while yielding interpretable mastery assessments. The adaptive recommendation strategy improves item relevance and personalization, and the LLM-based feedback offers targeted study guidance aligned with identified weaknesses. Overall, the results indicate that the proposed design is effective and practically deployable, providing a feasible pathway to generating individualized learning trajectories in intelligent education.
\end{abstract}


\section{Introduction}
As information technology advances, education is moving beyond one-size-fits-all instruction toward genuinely personalized learning \cite{Chen2025b,Wang2025b,Dong2025,Liao2024,Chen2024e,Wang2023v,Ma2023b,Wang2022as}. Conventional classroom models, however, still struggle to accommodate heterogeneity in students’ knowledge mastery, learning abilities, and study habits, limiting access to appropriate resources and guidance and, ultimately, dampening learning effectiveness \cite{Li2026a,Wang2024p,Li2023i,Wang2023j,Li2023g,Wang2025e}.

Cognitive Diagnosis Models (CDMs) and Computerized Adaptive Testing (CAT) offer principled tools to address this challenge by capturing individual differences and improving both assessment and instruction \cite{Zhang2024i,Wang2024b,Li2023f,Wang2023d}. CDMs provide fine-grained, interpretable estimates of students’ mastery at the knowledge-point level; recent deep-learning advances further enhance accuracy and diagnostic fidelity \cite{Wang2024s,Dong2023,Wang2023j}. CAT dynamically selects items based on real-time ability estimates, with maximum (Fisher/global) information criteria serving as classical and effective selection principles \cite{Dong2023,Li2023f}. When combined, CDMs’ granular posteriors and CAT’s adaptive mechanisms can power targeted resource recommendations and learning interventions that move beyond the traditional “one test for all” paradigm \cite{lou2023learning,Li2023g,Lyu2022}.

Despite steady progress, much of the prior work optimizes only one stage of this pipeline—modeling, item selection, or feedback—rather than delivering an integrated, closed-loop solution. This fragmentation often yields (i) coarse or weakly interpretable student models that overlook knowledge-point mastery, (ii) adaptive policies tied to restrictive parametric assumptions that underutilize diagnostic posteriors, and (iii) generic feedback that is not evidence-grounded and thus less actionable for learners.

Large Language Models (LLMs) introduce a complementary capability: converting diagnostic and recommendation outputs into fluent, structured, and actionable feedback. Compared with traditional methods, LLM-generated guidance can articulate specific weaknesses and next steps, helping students close gaps and refine strategies toward truly individualized instruction \cite{meyer2024etal,kinder2025etal,Chen2024g}.

To address these limitations, we propose and implement a personalized learning system that integrates Neural Cognitive Diagnosis (NeuralCD, NCD), a Bounded-Ability Estimation CAT strategy (BECAT), and LLM-based intelligent feedback. The NCD module performs fine-grained cognitive diagnosis at the knowledge-point level; BECAT uses bounded-ability estimation to select informative, level-appropriate items in real time; and the LLM component transforms diagnostic signals and recommendations into interpretable, evidence-grounded study guidance. Together, these components form a closed-loop framework of ``Diagnosis--Recommendation--Feedback,'' aligning modeling, measurement, and intervention within a single system.

This study makes three primary contributions: 

\begin{enumerate}
	\item We design an integrated architecture agent that unifies cognitive diagnosis, adaptive item selection, and feedback generation. 
	\item We develop a bounded-ability CAT policy that leverages diagnostic posteriors for stable, relevant recommendations.
	\item We introduce an LLM-based feedback module with a prompt-integration mechanism to deliver targeted, actionable guidance grounded in model evidence. We validate the system empirically on a public benchmark and analyze its effectiveness for personalized learning.
\end{enumerate}

The remainder of the paper is organized as follows. Section~\ref{RW} reviews related research on CDMs, adaptive testing and recommendation strategies, and educational applications of LLMs. Section~\ref{MTD} details the overall system workflow, the NCD model, the BECAT strategy, and the prompt-integration mechanism. Section~\ref{EXP} presents implementation details, datasets, and experimental results. Section~\ref{CON} concludes with limitations and future directions.

\begin{figure*}[htbp]
	\centerline{\includegraphics[width=\textwidth]{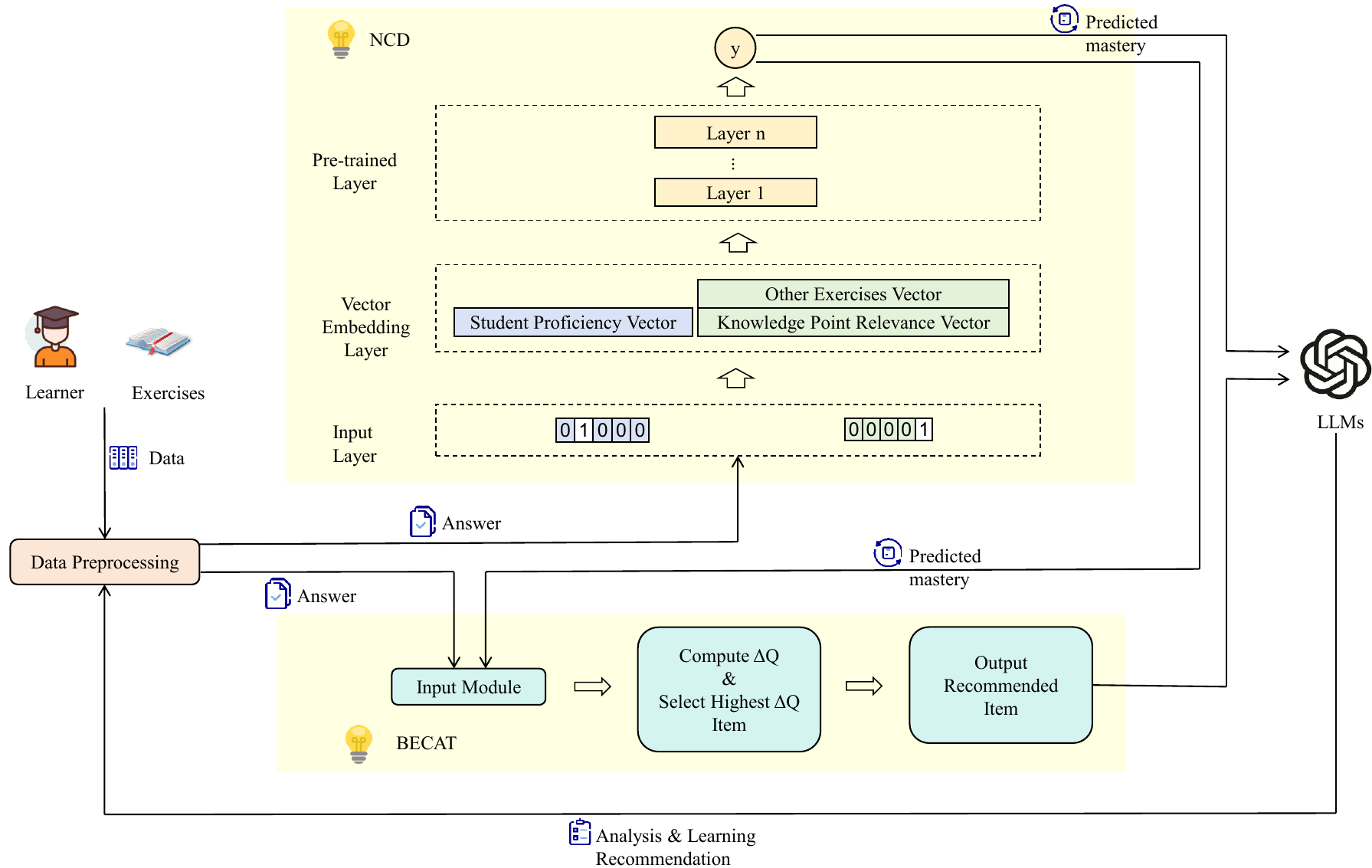}}
	\caption{The proposed EduLoop-Agent follows a four-stage workflow, forming a “Diagnosis–Recommendation–Feedback” closed loop.}
	\label{fig1}
\end{figure*}

\section{Related Work} \label{RW}
This section reviews three strands that underpin personalized learning: (i) cognitive diagnosis models, (ii) adaptive testing and recommendation strategies, and (iii) educational applications of large language models. We trace their evolution and synthesize common limitations—manual assumptions, heuristic or assumption-bound adaptivity, and stage-wise fragmentation—thereby motivating an integrated, closed-loop approach that aligns diagnosis, item selection, and feedback.

\subsection{Cognitive Diagnosis Models}
Since Rasch introduced Item Response Theory (IRT) in the 1960s~\cite{rasch1993probabilistic}, educational measurement has steadily advanced to better represent learners’ multidimensional knowledge and evolving skills. Key developments include Multidimensional IRT (MIRT)~\cite{reckase2009mirt}; cognitive diagnosis models (CDMs) grounded in Q-matrix specifications—such as DINA and G-DINA~\cite{delatorre2011gdina}; and Bayesian Knowledge Tracing (BKT) and its variants~\cite{corbett1995kt}. While influential, these approaches often depend on manually specified item–skill mappings, distributional assumptions, and parameter calibration, which can limit scalability and reduce fidelity in complex learning environments.

To mitigate these limitations based on deep learning \cite{Wang2025f,Zeng2025a,Wang2025,Zeng2024e,Wang2025d,Zeng2023c,Wang2024m,Li2023h,Wang2023g,Zeng2022,Wang2023v,Zeng2022b,Wang2023l,Zeng2021c,Wang2022ac,Zeng2020a,Wang2022at,Tian2018,Wang2021}, Wang et~al.\ proposed Neural Cognitive Diagnosis ~\cite{wang2020etal}, which leverages neural networks to learn student–item interactions while enforcing an attribute monotonicity assumption. NCD yields interpretable, fine-grained mastery estimates at the knowledge-point level and scales to large datasets. Beyond improved diagnostic accuracy, its skill-level posteriors naturally support downstream personalization, including targeted item recommendation to address weak prerequisites and the construction of coherent, prerequisite-aware learning paths.

\subsection{Adaptive Testing and Recommendation Strategies}
Adaptive testing and recommendation strategies are central to personalized learning systems: they aim to select, in real time, the most informative or pedagogically valuable next item based on a learner’s current state, thereby improving measurement efficiency and instructional precision. Early approaches largely relied on heuristic or parametric rules—e.g., maximum information criteria in IRT~\cite{rasch1993probabilistic} or mastery probabilities in BKT~\cite{corbett1995kt}. These methods typically require manual parameterization and Q-matrix specifications and can struggle with complex skill structures and cross-concept dependencies.

Recent work leverages deep learning \cite{Zheng2025,Wang2025g,Zeng2025,Chen2025a,Zeng2018,Chen2025,Zheng2024,Zeng2024g,Wang2020h,Zeng2024h,Wang2015b,Zeng2024b,Zhu2013,Zeng2024f,Wang2011,Zeng2024c,Zeng2024d,Zeng2024,Zeng2024a,Zeng2023a,Wang2018a,Zeng2023,Wang2023f,Chen2023b,Zeng2022a,Wang2022t,Zeng2021a,Zeng2020,Wang2011a} to overcome some of these limitations. Sequence models and attention mechanisms automatically extract latent features while capturing temporal learning behaviors and inter-skill relationships, leading to more dynamic and personalized recommendations~\cite{piech2015dkt,ghosh2020cak}. Graph-based approaches further incorporate knowledge graphs to guide item selection, promoting coherent learning paths and facilitating knowledge transfer~\cite{nakagawa2019gkt}. Overall, the field is moving toward more intelligent and fine-grained adaptive systems; however, many solutions still optimize item selection in isolation, without tightly integrating fine-grained diagnostic posteriors or closing the loop with evidence-grounded feedback—gaps addressed by the system proposed in this work.

\subsection{Applications of Large Language Models in Education}
Large language models (LLMs) have accelerated progress in intelligent education by enabling scalable, high-quality language generation. Transformer-based systems such as the GPT family can synthesize personalized learning materials and exercises aligned with learner profiles and trajectories, thereby supporting adaptive instruction and peer-like study resources. As interactive tutors, LLMs provide real-time formative feedback and clarification, improving the flexibility and accessibility of learning experiences~\cite{kinder2025etal}. In assessment, LLMs assist with automatic grading, item and hint generation, and learning analytics, which helps reduce instructor workload while maintaining evaluation coverage and timeliness~\cite{kasneci2023chatgpt}.

Despite these benefits, important challenges remain. LLM outputs can exhibit factual errors, calibration issues, and social biases; they are also sensitive to prompt design and may raise privacy and safety concerns in classroom deployments~\cite{nguyen2022etal}. Consequently, effective educational use typically pairs LLMs with grounding strategies (e.g., retrieval or tool integration), structured prompts aligned with instructional rubrics, human-in-the-loop review, and explicit safeguards. Building on these insights, our system employs LLMs not as standalone tutors but as feedback generators tethered to structured evidence: diagnostic posteriors from NCD and item selections from BECAT. This design aims to transform model decisions into interpretable, actionable guidance while mitigating generic or unsupported feedback.

\section{Proposed Method} \label{MTD}

EduLoop-Agent is an end-to-end, closed-loop personalized learning framework that operationalizes a ``Diagnosis--Recommendation--Feedback'' cycle as shown in Fig. \ref{fig1}. We first outline the workflow and then detail how NCD produces fine-grained mastery estimates, BECAT selects informative items in real time, and a prompt-integration mechanism grounds LLM feedback in model evidence to yield interpretable, actionable guidance.

\subsection{Workflow of Proposed EduLoop-Agent}
The proposed EduLoop-Agent follows a four-stage workflow, forming a “Diagnosis–Recommendation–Feedback” closed loop:
\subsubsection{Data Preprocessing}
The ASSISTments dataset is standardized to generate input files suitable for the NCD model and the BECAT strategy. These include an item–knowledge mapping table, student response records, a knowledge graph, and auxiliary mapping files (e.g., mapping between knowledge point IDs and textual descriptions), ensuring that both item information and knowledge structures are effectively utilized.
\subsubsection{Model Training}
Preprocessed data are fed into training scripts to train the NCD model and optimize its parameters. The outputs include trained model files and a table estimating students’ mastery levels across knowledge points, providing the basis for personalized recommendations.
\subsubsection{Adaptive Recommendation}
Leveraging the student mastery table and knowledge graph, data are input into the item recommendation scripts. The BECAT strategy dynamically selects items, enhancing the relevance and efficiency of recommendations while maintaining assessment accuracy.
\subsubsection{Personalized Feedback}
Student mastery data, recommended items, and relevant textual information are input into a Large Language Model (LLM), which generates interpretable feedback and actionable learning suggestions. This stage supports targeted interventions and the optimization of adaptive learning paths.

\subsection{Neural Cognitive Diagnosis Model}
The NCD model provides fine-grained and interpretable modeling of students’ knowledge mastery:
\subsubsection{Embedding and Feature Interaction}
Students, items, and knowledge points are mapped into low-dimensional vector spaces to form the ability vector $\vartheta_{s}\in\mathbb{R}^{d}$ of a student \textit{s}, the difficulty vector  $\beta_{q}\in\mathbb{R}^{d}$ of an item \textit{q} , and the discrimination scalar $\alpha_{q}\in\mathbb{R}^{+}$ of an item \textit{q}. The interaction between a student \textit{s} and an item \textit{q} is represented as:
\begin{equation}
    x_{s,q}=\alpha_{q}\cdot\quad(\vartheta_{s}-\beta_{q})
\end{equation}

This interaction feature captures the difference between student ability and item difficulty, modulated by the item discrimination, reflecting the relative impact of the item on the student’s performance.

\begin{figure}[htbp]
\centerline{\includegraphics[width=0.5\textwidth]{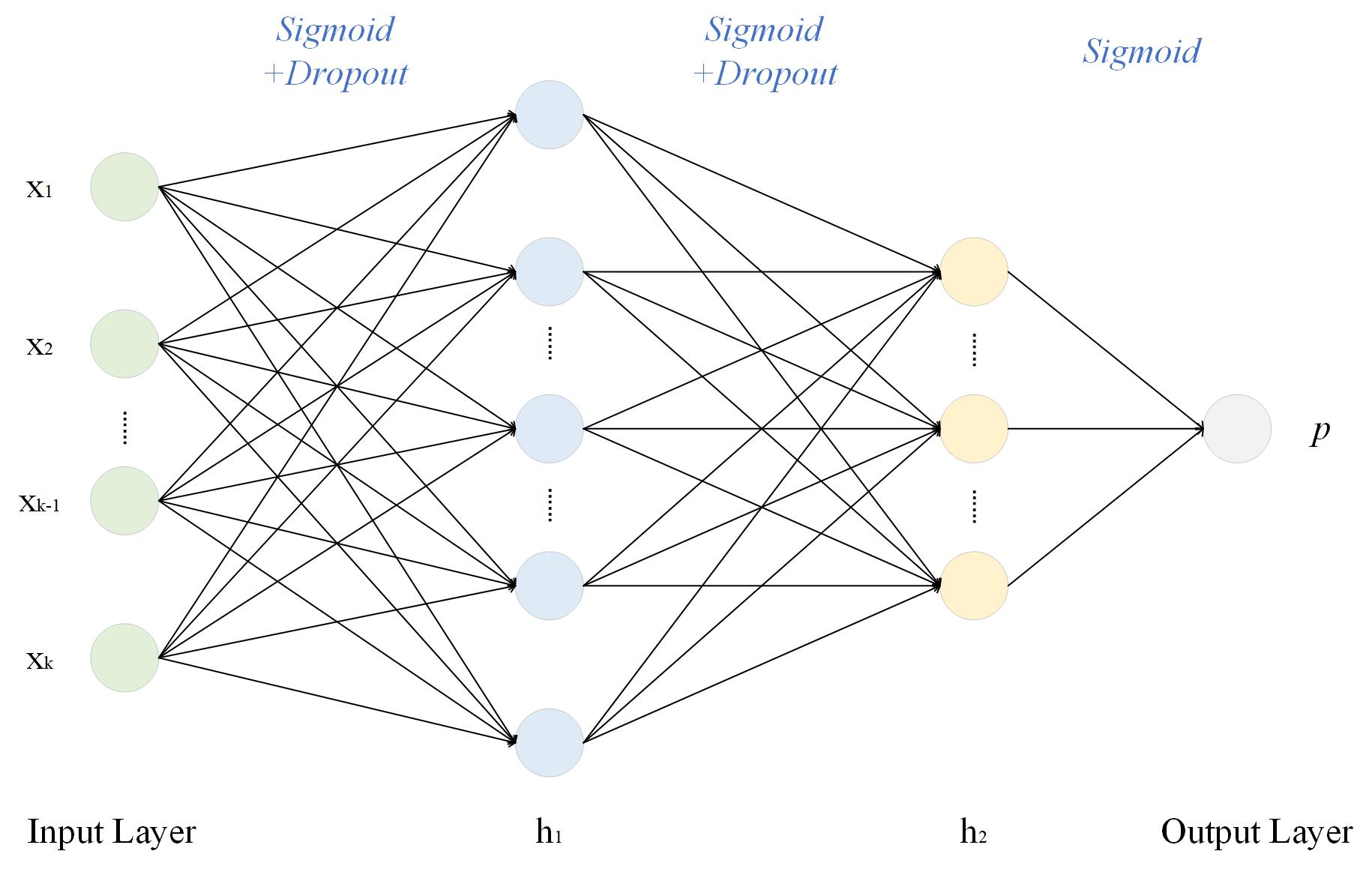}}
\caption{The embedding layer of the Neural Cognitive Diagnosis model.}
\label{fig2}
\end{figure}

\subsubsection{Prediction Network}
As shown in Fig. \ref{fig2}, the interaction feature is input into a multi-layer perceptron (\textit{MLP}), which maps it nonlinearly to the probability of a correct response:
\begin{equation}
    \hat{r}_{s,q}=\sigma(MLP(x_{s,q}))
\end{equation}
where $\sigma(\cdot)$ denotes the Sigmoid function, and the output $\hat{r}_{s,q}\in(0,1)$ represents the probability that student \textit{s} answers item \textit{q} correctly.
\subsubsection{Loss Function}
The model adopts binary cross-entropy (equivalent to negative log-likelihood) as the objective function:
\begin{equation}
    \mathscr{L}=-\frac{1}{N}\sum_{(s,q)}\left[r_{s,q}\mathrm{log}\hat{r}_{s,q}+(1-r_{s,q})\mathrm{log}(1-\hat{r}_{s,q})\right]
\end{equation}
where ${r}_{s,q}\in(0,1)$ represents the actual response. This loss quantifies the discrepancy between predicted and observed outcomes.

\subsection{Bounded Ability Estimation CAT Strategy}
The BECAT strategy dynamically selects items based on students’ estimated mastery levels:
\subsubsection{Expected Model Change (EMC)}
In an adaptive testing context, \textit{EMC} quantifies the contribution of each candidate item to updating student ability. Let $\Delta\theta^{(T)}$ and $\Delta\theta^{(F)}$ denote parameter changes under correct and incorrect responses. The expected model change for item \textit{q} is:
\begin{equation}
EMC(q)=\hat{r}_{s,q}\cdot\parallel\Delta\theta^{(T)}\parallel_2+\left(1-\hat{r}_{s,q}\right)\cdot\parallel\Delta\theta^{(F)}\parallel_2
\end{equation}
Higher \textit{EMC} values indicate greater potential to refine student ability estimates.
\subsubsection{Item Dependency Modeling}
A Bayesian weight matrix \textit{W} captures complementary information among items:
\begin{equation}
    W_{i,j}=C-\mathbb{E}[\parallel\nabla_\theta\ell_i-\nabla_\theta\ell_j\parallel]
\end{equation}
where $\ell_{i}$ and $\ell_{j}$ are the loss functions for items \textit{i} and \textit{j} , and \textit{C} ensures non-negativity. This matrix reflects inter-item dependencies and informs item selection.
\subsubsection{Information Score and Gain}
The information score \textit{F(S)} measures the coverage of unselected items relative to the selected set \textit{S}:
\begin{equation}
    F(S)=\sum_{i\notin S}\max_{j\in S}w_{ij}
\end{equation}
The information gain $\Delta_{\mathrm{q}}$ of candidate \textit{q} is defined as the score increment:
\begin{equation}
    \Delta_q=F(S\cup\{q\})-F(S)
\end{equation}
A greedy selection of items with maximal gain approximates an optimal combination.

\subsection{Prompt Integration Mechanism}
The LLM prompt template consists of fixed and dynamic components. The fixed component defines the LLM role (educational assessment expert and teaching assistant), specifies task requirements (analyzing mastery, evaluating recommendations, and providing learning suggestions), and sets output rules (structured bullet points with length limits). The dynamic component, populated automatically by the system, includes student mastery levels, recommended item texts, and corresponding knowledge points. Mastery levels, output by NCD, are presented as “Knowledge Point Name + Mastery Value,” while recommended items, output by BECAT, include item text and associated knowledge descriptions.

Output is required to be structured and actionable, with three sections: mastery analysis, recommendation evaluation, and personalized learning suggestions. This ensures relevance, readability, and effective integration of diagnosis, recommendation, and feedback into a coherent adaptive learning workflow.

\section{Experimental Results and Analysis} \label{EXP}
This section outlines the experimental setup, dataset preprocessing, and evaluation protocol used to assess EduLoop-Agent. We report results on the ASSISTments dataset, quantifying the NCD model’s predictive and diagnostic performance (AUC/ACC/RMSE/MSE/Loss) and visualizing training dynamics, then illustrate the downstream effects of BECAT-driven item selection and LLM-generated feedback through representative student cases.

\begin{figure}[htbp]
	\centerline{\includegraphics[width=0.5\textwidth]{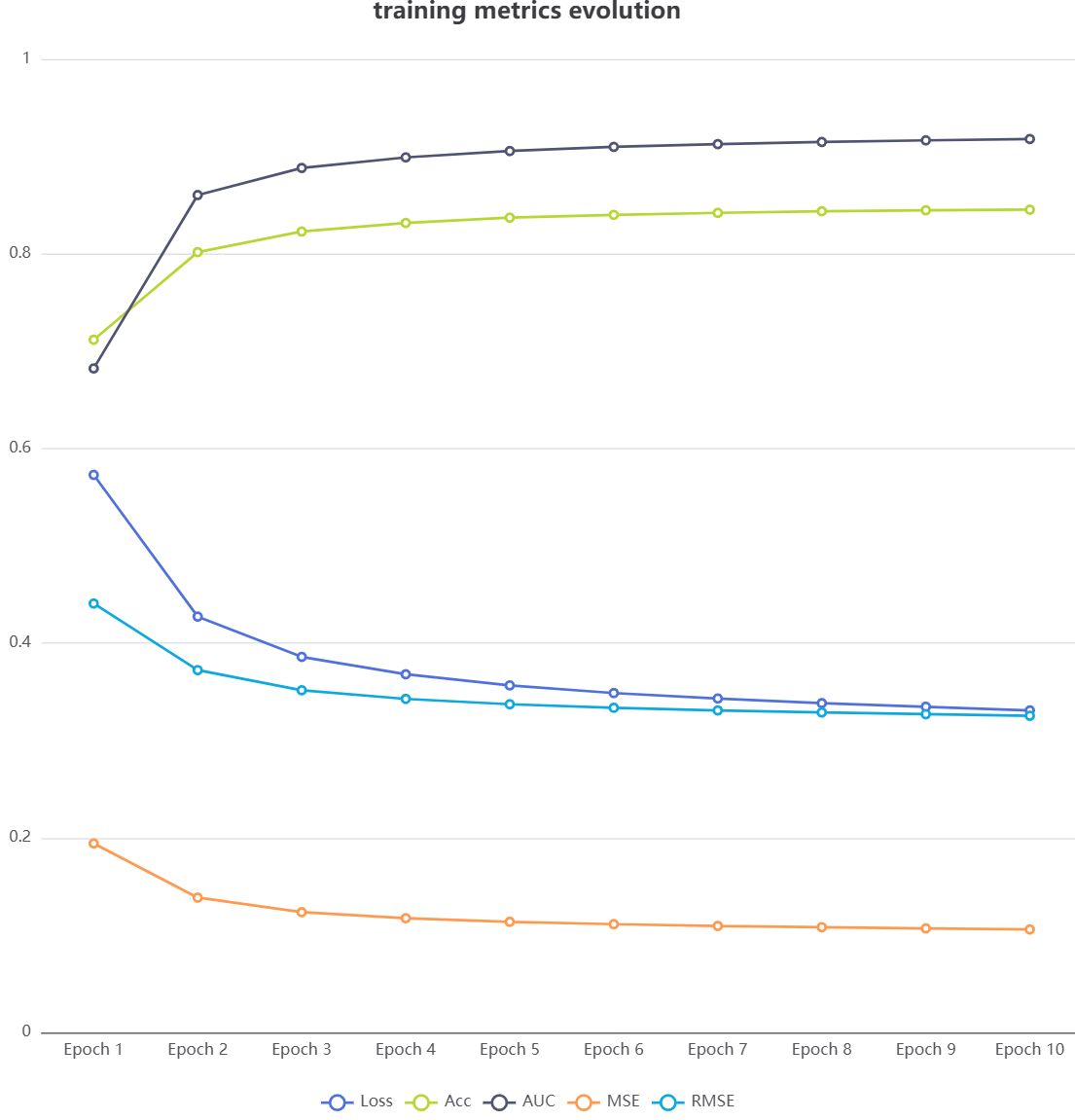}}
	\caption{The performance of the NCD model.}
	\label{fig3}
\end{figure}

\begin{figure}[htbp]
	\centerline{\includegraphics[width=0.5\textwidth]{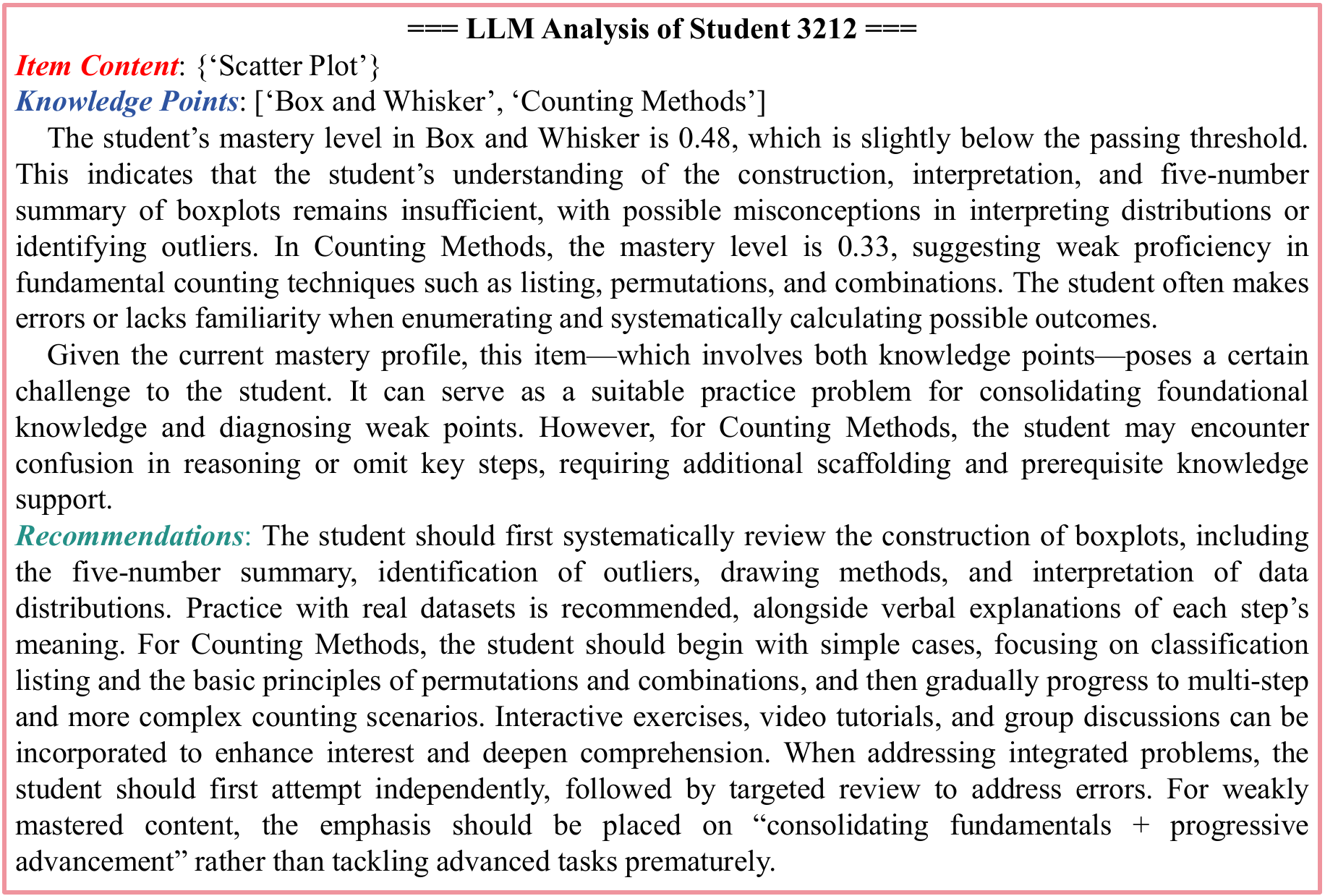}}
	\caption{EduLoop-Agent Generated Analysis Of Student 324.}
	\label{fig4}
\end{figure}

\begin{figure}[htbp]
	\centerline{\includegraphics[width=0.5\textwidth]{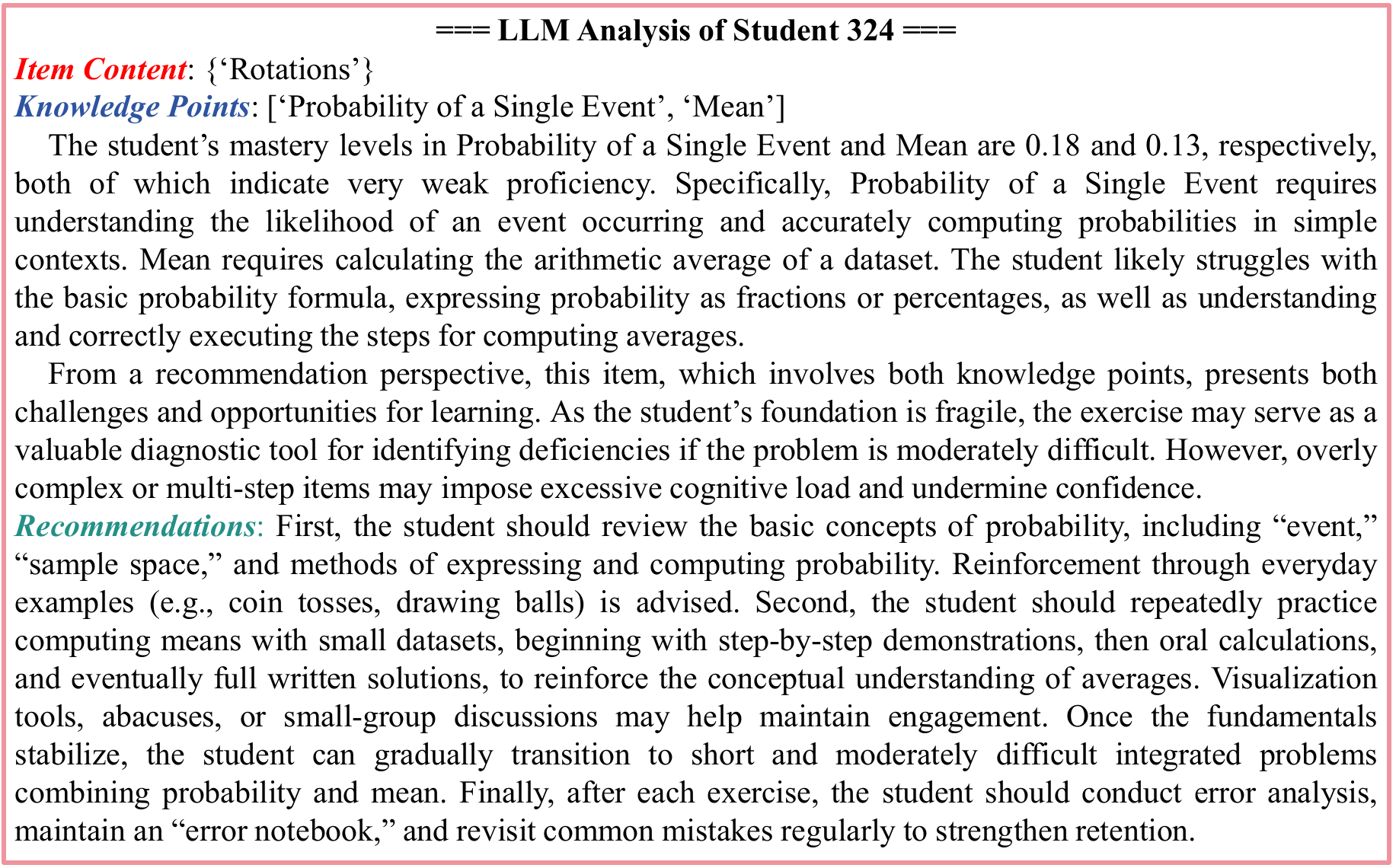}}
	\caption{EduLoop-Agent Generated Analysis Of Student 1718.}
	\label{fig5}
\end{figure}

\begin{figure}[htbp]
	\centerline{\includegraphics[width=0.5\textwidth]{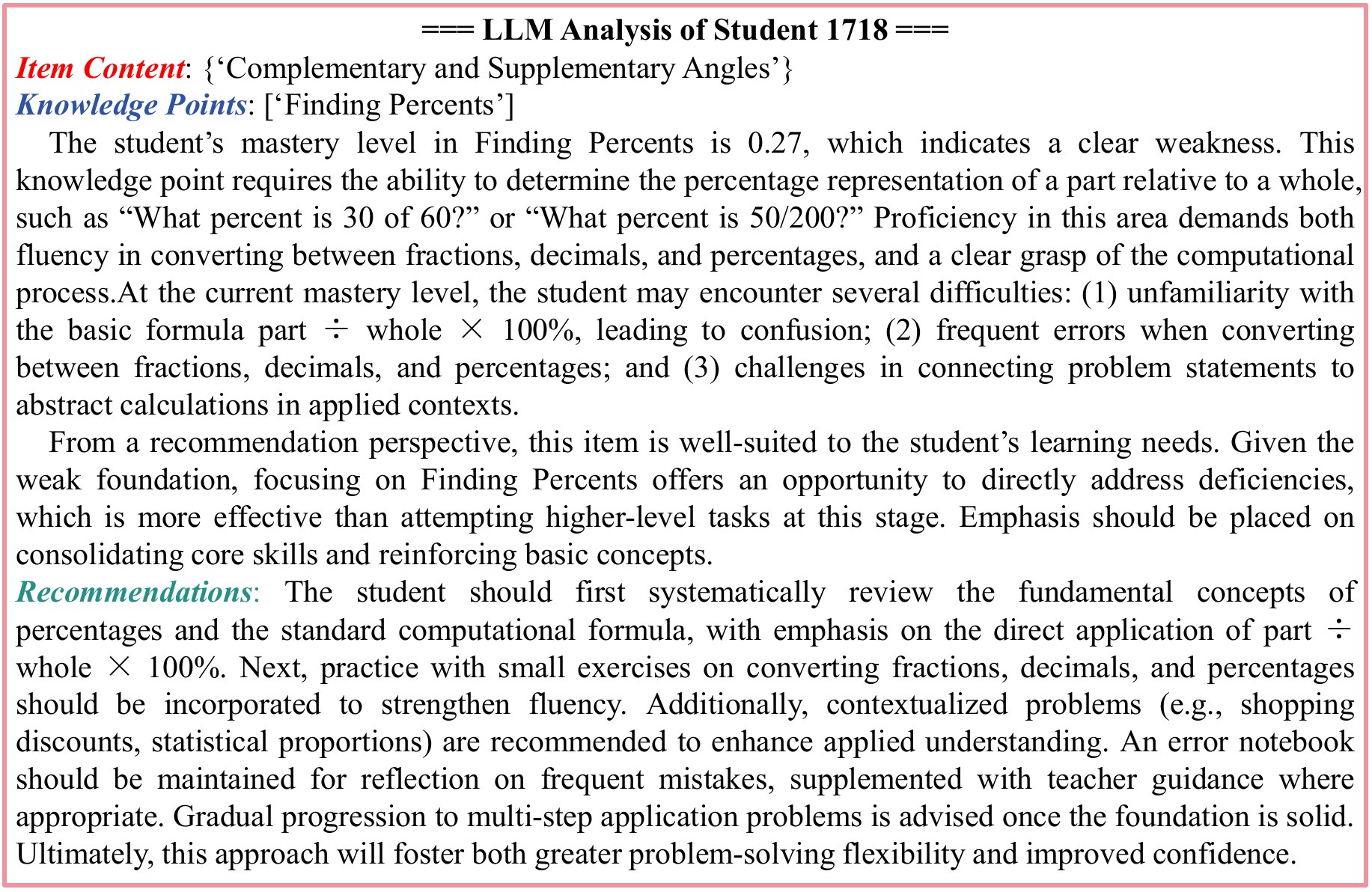}}
	\caption{EduLoop-Agent Generated Analysis Of Student 3212.}
	\label{fig6}
\end{figure}

\subsection{Experimental Setups}
The experimental environment of this study is based on Python 3.9, with the primary hardware being a computation server equipped with an NVIDIA GPU, running on Windows 11.
For tool selection, PyTorch is adopted as the core deep learning framework, responsible for the training and inference of the Neural Cognitive Diagnosis (NCD) model. Meanwhile, the transformers library is used to load and invoke large language model (LLM) interfaces, enabling prompt-based feedback generation and natural language processing. To implement the adaptive recommendation strategy, the CAT library is employed to realize the Bounded Ability Estimation Computerized Adaptive Testing (BECAT) strategy, which is integrated with the NCD module.

In the feedback generation stage, the LLM API is called to integrate cognitive diagnostic results with recommended item information, which are then input into the model to produce personalized instructional feedback.

\subsection{Dataset}
The experiments are conducted on the publicly available ASSISTments dataset, which has been widely used in knowledge tracing and adaptive testing research, offering strong generalizability and comparability. The dataset contains student response records, item–knowledge point annotations, and correctness labels (correct/incorrect). During preprocessing, the raw response logs are cleaned and reformatted, ensuring consistency in the mapping of student IDs, item IDs, and knowledge point IDs.

\subsection{Result Analysis}
\subsubsection{Performance of the NCD Model}
Based on the NCD model, this study establishes a cognitive diagnosis framework to predict both the probabilities of students correctly answering items and their levels of mastery over knowledge points. The predictions are compared with the actual response data and the evaluation metrics, including AUC, ACC, RMSE, MSE, and LOSS are calculated. The experimental results are visualized using Echarts software.

As shown in Fig. \ref{fig3}, the results demonstrate that model training is stable: the loss metric decreases smoothly without significant oscillations, indicating a normal optimization process. Both ACC and AUC show notable improvement, with AUC exceeding 0.9, reflecting strong discriminative capability. The continuous decline of MSE and RMSE indicates that the deviation between predicted and actual values is steadily reduced, highlighting the increasing accuracy of the model. Overall, after 10 epochs of training, the NCD model demonstrates strong performance: it not only predicts student responses accurately but also provides reliable estimates of mastery levels, making it well-suited for personalized learning diagnosis and recommendation.

\subsubsection{Item Recommendation Results and Feedback}
For illustrative purposes, we randomly selected three students to demonstrate the analysis results as shown in Fig. \ref{fig4}, \ref{fig5} and \ref{fig6}:

\section{Conclusions} \label{CON}
This paper presented EduLoop-Agent, an end-to-end, closed-loop personalized learning framework that unifies Neural Cognitive Diagnosis, a bounded-ability CAT strategy, and LLM-based feedback to align diagnosis, adaptive recommendation, and intervention within a single “Diagnosis--Recommendation--Feedback” cycle. Experiments on the ASSISTments dataset indicate stable training dynamics, strong response-prediction performance, improved recommendation relevance, and feedback that is interpretable and pedagogically actionable. Nonetheless, the study’s external validity is limited by reliance on a single public dataset and offline simulations, and LLM-generated feedback can be sensitive to prompt design, exhibit variability, and inherit social biases, highlighting the need for better calibration and factual grounding. Future work will broaden evaluation across subjects, grade levels, and institutions; include ablations and user studies with teachers and students; enhance reliability via retrieval-grounded prompting, constrained decoding, rubric-aligned evaluation, and human-in-the-loop review; and conduct classroom deployments to measure learning gains, recommendation efficiency, and fairness under explicit privacy and safety safeguards.

\bibliographystyle{IEEEtran}
\bibliography{mine,Citations}

\begin{thebibliography}{10}
\providecommand{\url}[1]{#1}
\csname url@samestyle\endcsname
\providecommand{\newblock}{\relax}
\providecommand{\bibinfo}[2]{#2}
\providecommand{\BIBentrySTDinterwordspacing}{\spaceskip=0pt\relax}
\providecommand{\BIBentryALTinterwordstretchfactor}{4}
\providecommand{\BIBentryALTinterwordspacing}{\spaceskip=\fontdimen2\font plus
\BIBentryALTinterwordstretchfactor\fontdimen3\font minus
  \fontdimen4\font\relax}
\providecommand{\BIBforeignlanguage}[2]{{%
\expandafter\ifx\csname l@#1\endcsname\relax
\typeout{** WARNING: IEEEtran.bst: No hyphenation pattern has been}%
\typeout{** loaded for the language `#1'. Using the pattern for}%
\typeout{** the default language instead.}%
\else
\language=\csname l@#1\endcsname
\fi
#2}}
\providecommand{\BIBdecl}{\relax}
\BIBdecl

\bibitem{Chen2025b}
Z.~Chen, D.~Ren, Z.~Wang, L.~Gao, Y.~Shi, H.~Liu, and T.~Yang, ``Enhancing
  student academic performance in middle school classrooms by fostering
  engagement motivation through intelligent assessment of teacher praise
  emotional intensity,'' \emph{Learning and Instruction}, vol.~99, p. 102185,
  Oct. 2025.

\bibitem{Wang2025b}
Y.~Wang, M.~Zuo, X.~He, and Z.~Wang, ``Exploring students online learning
  behavioral engagement in university: Factors, academic performance and their
  relationship,'' \emph{Behavioral Sciences}, vol.~15, no.~1, p.~78, Jan. 2025.

\bibitem{Dong2025}
S.~Dong, X.~Niu, R.~Zhong, Z.~Wang, and M.~Zuo, ``Leveraging label semantics
  and meta-label refinement for multi-label question classification,''
  \emph{Knowledge-Based Systems}, vol. 318, p. 113412, Jun. 2025.

\bibitem{Liao2024}
X.~Liao, X.~Zhang, Z.~Wang, and H.~Luo, ``Design and implementation of an
  ai-enabled visual report tool as formative assessment to promote learning
  achievement and self-regulated learning: An experimental study,''
  \emph{British Journal of Educational Technology}, vol.~55, no.~3, pp.
  1253--1276, 2024.

\bibitem{Chen2024e}
D.~Chen, X.~Xu, Z.~Wang, and J.~Shen, ``Personalized art image generation model
  for smart art education,'' in \emph{2024 International Conference on
  Intelligent Education and Intelligent Research (IEIR)}.\hskip 1em plus 0.5em
  minus 0.4em\relax Macau, China: IEEE, Nov. 2024, pp. 1--8.

\bibitem{Wang2023v}
Z.~Wang, M.~Wang, C.~Zeng, J.~Yao, Y.~Yang, and H.~Xu, ``Enhanced convolutional
  neural networks based learner authentication for personalized e-learning
  system,'' in \emph{2023 International Conference on Intelligent Education and
  Intelligent Research (IEIR)}.\hskip 1em plus 0.5em minus 0.4em\relax Wuhan,
  China: IEEE, Nov. 2023, pp. 1--7.

\bibitem{Ma2023b}
L.~Ma, X.~Zhang, Z.~Wang, and H.~Luo, ``Designing effective instructional
  feedback using a diagnostic and visualization system: Evidence from a high
  school biology class,'' \emph{Systems}, vol.~11, no.~7, p. 364, Jul. 2023.

\bibitem{Wang2022as}
Z.~Wang, W.~Wu, C.~Zeng, J.~Yao, Y.~Yang, and H.~Xu, ``Smart contract
  vulnerability detection for educational blockchain based on graph neural
  networks,'' in \emph{2022 International Conference on Intelligent Education
  and Intelligent Research (IEIR)}.\hskip 1em plus 0.5em minus 0.4em\relax
  IEEE, Dec. 2022, pp. 8--14.

\bibitem{Li2026a}
L.~Li, Z.~Wang, J.~M. Jose, and X.~Ge, ``Llm supporting knowledge tracing
  leveraging global subject and student specific knowledge graphs,''
  \emph{Information Fusion}, vol. 126, p. 103577, Feb. 2026.

\bibitem{Wang2024p}
Z.~Wang, M.~Su, Y.~Yang, C.~Zeng, and L.~Ye, ``Cross-disciplinary cognitive
  diagnosis leveraging deep transfer learning for smart education,'' in
  \emph{2024 International Conference on Intelligent Education and Intelligent
  Research (IEIR)}.\hskip 1em plus 0.5em minus 0.4em\relax Macau, China: IEEE,
  Nov. 2024, pp. 1--8.

\bibitem{Li2023i}
L.~Li and Z.~Wang, ``Knowledge relation rank enhanced heterogeneous learning
  interaction modeling for neural graph forgetting knowledge tracing,''
  \emph{PLOS ONE}, vol.~18, no.~12, p. e0295808, Dec. 2023.

\bibitem{Wang2023j}
Z.~Wang, W.~Yan, C.~Zeng, Y.~Tian, and S.~Dong, ``A unified interpretable
  intelligent learning diagnosis framework for learning performance prediction
  in intelligent tutoring systems,'' \emph{International Journal of Intelligent
  Systems}, vol. 2023, p. e4468025, Feb. 2023.

\bibitem{Li2023g}
L.~Li and Z.~Wang, ``Knowledge graph-enhanced intelligent tutoring system based
  on exercise representativeness and informativeness,'' \emph{International
  Journal of Intelligent Systems}, vol. 2023, p. e2578286, Oct. 2023.

\bibitem{Wang2025e}
Z.~Wang, Z.~Lu, C.~Zeng, S.~Dong, M.~Zuo, and J.~Sun, ``Mmkt: Multimodal
  knowledge tracing in personalized e-learning systems for supporting lifelong
  learning,'' \emph{IEEE Transactions on Computational Social Systems}, pp.
  1--20, 2025.

\bibitem{Zhang2024i}
Y.~Zhang, Y.~Zhang, W.~Xu, Z.~Wang, and J.~Sun, ``Singpad: A knowledge tracing
  dataset based on music performance assessment,'' in \emph{Proceedings of the
  17th International Conference on Educational Data Mining}, 2024, pp.
  332--340.

\bibitem{Wang2024b}
Z.~Wang, W.~Wu, C.~Zeng, H.~Luo, and J.~Sun, ``Psychological factors enhanced
  heterogeneous learning interactive graph knowledge tracing for understanding
  the learning process,'' \emph{Frontiers in Psychology}, vol.~15, May 2024.

\bibitem{Li2023f}
L.~Li and Z.~Wang, ``Calibrated q-matrix-enhanced deep knowledge tracing with
  relational attention mechanism,'' \emph{Applied Sciences}, vol.~13, no.~4,
  pp. 1--24, Jan. 2023.

\bibitem{Wang2023d}
Z.~Wang, Y.~Hou, C.~Zeng, S.~Zhang, and R.~Ye, ``Multiple learning
  features--enhanced knowledge tracing based on learner--resource response
  channels,'' \emph{Sustainability}, vol.~15, no.~12, p. 9427, Jan. 2023.

\bibitem{Wang2024s}
Z.~Wang, J.~Wan, Y.~Yang, C.~Zeng, and J.~Shen, ``Unified knowledge tracing
  framework for subjective and objective assessments,'' in \emph{2024
  International Conference on Intelligent Education and Intelligent Research
  (IEIR)}.\hskip 1em plus 0.5em minus 0.4em\relax Macau, China: IEEE, Nov.
  2024, pp. 1--8.

\bibitem{Dong2023}
S.~Dong, X.~Tao, R.~Zhong, Z.~Wang, M.~Zuo, and J.~Sun, ``Advanced mathematics
  exercise recommendation based on automatic knowledge extraction and
  multi-layer knowledge graph,'' \emph{IEEE Transactions on Learning
  Technologies}, pp. 1--17, 2023.

\bibitem{lou2023learning}
P.~Lou, ``Learning path recommendation of intelligent education based on
  cognitive diagnosis,'' \emph{International Journal of Emerging Technologies
  in Learning}, vol.~18, no.~13, 2023.

\bibitem{Lyu2022}
L.~Lyu, Z.~Wang, H.~Yun, Z.~Yang, and Y.~Li, ``Deep knowledge tracing based on
  spatial and temporal representation learning for learning performance
  prediction,'' \emph{Applied Sciences}, vol.~12, no.~14, pp. 1--21, Jan. 2022.

\bibitem{meyer2024etal}
J.~Meyer, T.~Jansen, R.~Schiller, L.~W. Liebenow, M.~Steinbach, A.~Horbach, and
  J.~Fleckenstein, ``Using llms to bring evidence-based feedback into the
  classroom: Ai-generated feedback increases secondary students’ text
  revision, motivation, and positive emotions,'' \emph{Computers and Education:
  Artificial Intelligence}, vol.~6, p. 100199, 2024.

\bibitem{kinder2025etal}
A.~Kinder, F.~J. Briese, M.~Jacobs, N.~Dern, N.~Glodny, S.~Jacobs, and
  S.~Le{\ss}mann, ``Effects of adaptive feedback generated by a large language
  model: A case study in teacher education,'' \emph{Computers and Education:
  Artificial Intelligence}, vol.~8, p. 100349, 2025.

\bibitem{Chen2024g}
Z.~Chen, L.~Gao, Z.~Wang, D.~Ren, T.~Yang, and Y.~Shi, ``Enhancing large
  language models for precise classification of teacher praise discourse: A
  fine-tuning approach,'' in \emph{2024 International Conference on Intelligent
  Education and Intelligent Research (IEIR)}.\hskip 1em plus 0.5em minus
  0.4em\relax Macau, China: IEEE, Nov. 2024, pp. 1--6.

\bibitem{rasch1993probabilistic}
G.~Rasch, \emph{Probabilistic Models for Some Intelligence and Attainment
  Tests}, expanded~ed.\hskip 1em plus 0.5em minus 0.4em\relax Chicago, IL:
  University of Chicago Press, 1993.

\bibitem{reckase2009mirt}
M.~D. Reckase, ``Multidimensional item response theory models,'' in
  \emph{Multidimensional Item Response Theory}.\hskip 1em plus 0.5em minus
  0.4em\relax New York, NY: Springer, 2009, pp. 79--112.

\bibitem{delatorre2011gdina}
J.~De~La~Torre, ``The generalized dina model framework,'' \emph{Psychometrika},
  vol.~76, no.~2, pp. 179--199, 2011.

\bibitem{corbett1995kt}
A.~T. Corbett and J.~R. Anderson, ``Knowledge tracing: Modeling the acquisition
  of procedural knowledge,'' \emph{User Modeling and User-Adapted Interaction},
  vol.~4, no.~4, pp. 253--278, 1995.

\bibitem{Wang2025f}
Z.~Wang, M.~Wang, C.~Zeng, and L.~Li, ``Scb-detr: Multiscale deformable
  transformers for occlusion-resilient student learning behavior detection in
  smart classroom,'' \emph{IEEE Transactions on Computational Social Systems},
  pp. 1--20, 2025.

\bibitem{Zeng2025a}
C.~Zeng, F.~Zou, S.~Xia, and Z.~Wang, ``Sau-net: Saliency-based adaptive
  unfolding network for interpretable high-quality image compressed sensing in
  internet of things,'' \emph{IEEE Internet of Things Journal}, vol.~12,
  no.~14, pp. 28\,672--28\,688, Jul. 2025.

\bibitem{Wang2025}
Z.~Wang, L.~Li, C.~Zeng, S.~Dong, and J.~Sun, ``Slbdetection-net: Towards
  closed-set and open-set student learning behavior detection in smart
  classroom of k-12 education,'' \emph{Expert Systems with Applications}, vol.
  260, p. 125392, Jan. 2025.

\bibitem{Zeng2024e}
C.~Zeng, Y.~Yu, Z.~Wang, S.~Xia, H.~Cui, and X.~Wan, ``Gsista-net: Generalized
  structure ista networks for image compressed sensing based on optimized
  unrolling algorithm,'' \emph{Multimedia Tools and Applications}, Mar. 2024.

\bibitem{Wang2025d}
Z.~Wang, L.~Li, C.~Zeng, S.~Dong, and J.~Sun, ``Slb-mamba: A vision mamba for
  closed and open-set student learning behavior detection,'' \emph{Applied Soft
  Computing}, vol. 180, p. 113369, Aug. 2025.

\bibitem{Zeng2023c}
C.~Zeng, S.~Xia, Z.~Wang, and X.~Wan, ``Multi-channel representation learning
  enhanced unfolding multi-scale compressed sensing network for high quality
  image reconstruction,'' \emph{Entropy}, vol.~25, no.~12, p. 1579, Dec. 2023.

\bibitem{Wang2024m}
Z.~Wang, M.~Wang, C.~Zeng, and L.~Li, ``Sbd-net: Incorporating multi-level
  features for an efficient detection network of student behavior in smart
  classrooms,'' \emph{Applied Sciences}, vol.~14, no.~18, p. 8357, Jan. 2024.

\bibitem{Li2023h}
L.~Li, Z.~Wang, and T.~Zhang, ``Gbh-yolov5: Ghost convolution with
  bottleneckcsp and tiny target prediction head incorporating yolov5 for pv
  panel defect detection,'' \emph{Electronics}, vol.~12, no.~3, pp. 1--15, Jan.
  2023.

\bibitem{Wang2023g}
Z.~Wang, J.~Yao, C.~Zeng, L.~Li, and C.~Tan, ``Students' classroom behavior
  detection system incorporating deformable detr with swin transformer and
  light-weight feature pyramid network,'' \emph{Systems}, vol.~11, no.~7, p.
  372, Jul. 2023.

\bibitem{Zeng2022}
C.~Zeng, J.~Ye, Z.~Wang, N.~Zhao, and M.~Wu, ``Cascade neural network-based
  joint sampling and reconstruction for image compressed sensing,''
  \emph{Signal, Image and Video Processing}, vol.~16, no.~1, pp. 47--54, Feb.
  2022.

\bibitem{Zeng2022b}
C.~Zeng, K.~Yan, Z.~Wang, Y.~Yu, S.~Xia, and N.~Zhao, ``Abs-cam: A gradient
  optimization interpretable approach for explanation of convolutional neural
  networks,'' \emph{Signal, Image and Video Processing}, pp. 1--8, Jul. 2022.

\bibitem{Wang2023l}
Z.~Wang, L.~Li, C.~Zeng, and J.~Yao, ``Student learning behavior recognition
  incorporating data augmentation with learning feature representation in smart
  classrooms,'' \emph{Sensors}, vol.~23, no.~19, p. 8190, Jan. 2023.

\bibitem{Zeng2021c}
C.~Zeng, Z.~Wang, Z.~Wang, K.~Yan, and Y.~Yu, ``Image compressed sensing and
  reconstruction of multi-scale residual network combined with channel
  attention mechanism,'' \emph{Journal of Physics: Conference Series}, vol.
  2010, no.~1, p. 012134, Sep. 2021.

\bibitem{Wang2022ac}
Z.~Wang, Z.~Wang, C.~Zeng, Y.~Yu, and X.~Wan, ``High-quality image compressed
  sensing and reconstruction with multi-scale dilated convolutional neural
  network,'' \emph{Circuits, Systems, and Signal Processing}, pp. 1--24, Sep.
  2022.

\bibitem{Zeng2020a}
C.~Zeng, Z.~Wang, and Z.~Wang, ``Image reconstruction of iot based on parallel
  cnn,'' in \emph{2020 International Conferences on Internet of Things
  (iThings)}.\hskip 1em plus 0.5em minus 0.4em\relax Rhodes, Greece: IEEE, Nov.
  2020, pp. 258--263.

\bibitem{Wang2022at}
Z.~Wang, J.~Yao, C.~Zeng, W.~Wu, H.~Xu, and Y.~Yang, ``Yolov5 enhanced learning
  behavior recognition and analysis in smart classroom with multiple
  students,'' in \emph{2022 International Conference on Intelligent Education
  and Intelligent Research (IEIR)}.\hskip 1em plus 0.5em minus 0.4em\relax
  IEEE, Dec. 2022, pp. 23--29.

\bibitem{Tian2018}
Y.~Tian, X.~Wang, H.~Yao, J.~Chen, Z.~Wang, and L.~Yi, ``Occlusion handling
  using moving volume and ray casting techniques for augmented reality
  systems,'' \emph{Multimedia Tools and Applications}, vol.~77, no.~13, pp.
  16\,561--16\,578, Jul. 2018.

\bibitem{Wang2021}
Z.~Wang, C.~Zuo, and C.~Zeng, ``Sae based unified double jpeg compression
  detection system for web image forensics,'' \emph{International Journal of
  Web Information Systems}, vol.~17, no.~2, pp. 84--98, Apr. 2021.

\bibitem{wang2020etal}
F.~Wang, Q.~Liu, E.~Chen \emph{et~al.}, ``Neural cognitive diagnosis for
  intelligent education systems,'' in \emph{AAAI}, 2020.

\bibitem{Zheng2025}
Q.~Zheng, Z.~Chen, Z.~Wang, G.~Hou, C.~Liu, C.~Zhao, and C.~Zou, ``Enhancing
  domain adaptation in speaker verification via partially shared adversarial
  network,'' in \emph{Advanced Intelligent Computing Technology and
  Applications}, D.-S. Huang, Q.~Zhang, C.~Zhang, and W.~Chen, Eds.\hskip 1em
  plus 0.5em minus 0.4em\relax Singapore: Springer Nature, 2025, pp. 158--168.

\bibitem{Wang2025g}
Z.~Wang, L.~Li, and C.~Zeng, ``Dwmgrad: An innovative neural network
  optimization approach using dynamic window data for adaptive updating of
  momentum and learning rate,'' \emph{Applied Intelligence}, vol.~55, no.~15,
  p. 1026, Oct. 2025.

\bibitem{Zeng2025}
C.~Zeng, Y.~Zhao, and Z.~Wang, ``Smdrl: Self-supervised mobile device
  representation learning framework for recording source identification from
  unlabeled data,'' \emph{Expert Systems with Applications}, vol. 282, p.
  127635, Jul. 2025.

\bibitem{Chen2025a}
Z.~Chen, C.~Zhao, Z.~Wang, C.~Liu, Q.~Zheng, and C.~Zou, ``Ds-btian: A novel
  deep-shallow bidirectional transformer interactive attention network for
  multimodal emotion recognition,'' in \emph{ICASSP 2025 - 2025 IEEE
  International Conference on Acoustics, Speech and Signal Processing
  (ICASSP)}, Apr. 2025, pp. 1--5.

\bibitem{Zeng2018}
C.-Y. Zeng, C.-F. Ma, Z.-F. Wang, and J.-X. Ye, ``Stacked autoencoder networks
  based speaker recognition,'' in \emph{2018 International Conference on
  Machine Learning and Cybernetics (ICMLC)}.\hskip 1em plus 0.5em minus
  0.4em\relax Chengdu: IEEE, Jul. 2018, pp. 294--299.

\bibitem{Chen2025}
Z.~Chen, C.~Liu, Z.~Wang, C.~Zhao, M.~Lin, and Q.~Zheng, ``Mtlser: Multi-task
  learning enhanced speech emotion recognition with pre-trained acoustic
  model,'' \emph{Expert Systems with Applications}, vol. 273, p. 126855, May
  2025.

\bibitem{Zheng2024}
Q.~Zheng, Z.~Chen, Z.~Wang, H.~Liu, and M.~Lin, ``Meconformer: Highly
  representative embedding extractor for speaker verification via incorporating
  selective convolution into deep speaker encoder,'' \emph{Expert Systems with
  Applications}, vol. 244, p. 123004, Jun. 2024.

\bibitem{Zeng2024g}
C.~Zeng, Y.~Zhao, Z.~Wang, K.~Li, X.~Wan, and M.~Liu, ``Squeeze-and-excitation
  self-attention mechanism enhanced digital audio source recognition based on
  transfer learning,'' \emph{Circuits, Systems, and Signal Processing}, Sep.
  2024.

\bibitem{Wang2020h}
Z.~Wang, S.~Duan, C.~Zeng, X.~Yu, Y.~Yang, and H.~Wu, ``Robust speaker
  identification of iot based on stacked sparse denoising auto-encoders,'' in
  \emph{2020 International Conferences on Internet of Things (iThings)}.\hskip
  1em plus 0.5em minus 0.4em\relax Rhodes, Greece: IEEE, Nov. 2020, pp.
  252--257.

\bibitem{Zeng2024h}
C.~Zeng, Y.~Zhao, and Z.~Wang, ``Mobileformer: Cross-scale and multi-level
  representation learning based mobile recording device recognition for
  consumer electronics,'' \emph{IEEE Transactions on Consumer Electronics}, pp.
  1--1, 2024.

\bibitem{Wang2015b}
Z.~Wang, Q.~Liu, J.~Chen, and H.~Yao, ``Recording source identification using
  device universal background model,'' in \emph{2015 International Conference
  of Educational Innovation through Technology (EITT)}.\hskip 1em plus 0.5em
  minus 0.4em\relax Wuhan, China: IEEE, Oct. 2015, pp. 19--23.

\bibitem{Zeng2024b}
C.~Zeng, K.~Li, and Z.~Wang, ``Enfformer: Long-short term representation of
  electric network frequency for digital audio tampering detection,''
  \emph{Knowledge-Based Systems}, vol. 297, p. 111938, Aug. 2024.

\bibitem{Zhu2013}
Z.-Y. Zhu, Q.-H. He, X.-H. Feng, Y.-X. Li, and Z.-F. Wang, ``Liveness detection
  using time drift between lip movement and voice,'' in \emph{2013
  International Conference on Machine Learning and Cybernetics}, vol.~02, Jul.
  2013, pp. 973--978.

\bibitem{Zeng2024f}
C.~Zeng, S.~Kong, Z.~Wang, K.~Li, Y.~Zhao, X.~Wan, and Y.~Chen,
  ``Discriminative component analysis enhanced feature fusion of electrical
  network frequency for digital audio tampering detection,'' \emph{Circuits,
  Systems, and Signal Processing}, Jul. 2024.

\bibitem{Wang2011}
Z.-F. Wang, G.~Wei, and Q.-H. He, ``Channel pattern noise based playback attack
  detection algorithm for speaker recognition,'' in \emph{2011 International
  Conference on Machine Learning and Cybernetics}, vol.~4, Jul. 2011, pp.
  1708--1713.

\bibitem{Zeng2024c}
C.~Zeng, S.~Kong, Z.~Wang, K.~Li, Y.~Zhao, X.~Wan, and Y.~Chen, ``Digital audio
  tampering detection based on spatio-temporal representation learning of
  electrical network frequency,'' \emph{Multimedia Tools and Applications}, pp.
  1--21, Mar. 2024.

\bibitem{Zeng2024d}
C.~Zeng, S.~Kong, Z.~Wang, S.~Feng, N.~Zhao, and J.~Wang, ``Deletion and
  insertion tampering detection for speech authentication based on fluctuating
  super vector of electrical network frequency,'' \emph{Speech Communication},
  vol. 158, p. 103046, Mar. 2024.

\bibitem{Zeng2024}
C.~Zeng, S.~Feng, Z.~Wang, X.~Wan, Y.~Chen, and N.~Zhao, ``Spatio-temporal
  representation learning enhanced source cell-phone recognition from speech
  recordings,'' \emph{Journal of Information Security and Applications},
  vol.~80, p. 103672, Feb. 2024.

\bibitem{Zeng2024a}
C.~Zeng, S.~Feng, Z.~Wang, Y.~Zhao, K.~Li, and X.~Wan, ``Audio source recording
  device recognition based on representation learning of sequential gaussian
  mean matrix,'' \emph{Forensic Science International: Digital Investigation},
  vol.~48, p. 301676, Mar. 2024.

\bibitem{Zeng2023a}
C.~Zeng, S.~Kong, Z.~Wang, K.~Li, and Y.~Zhao, ``Digital audio tampering
  detection based on deep temporal--spatial features of electrical network
  frequency,'' \emph{Information}, vol.~14, no.~5, p. 253, May 2023.

\bibitem{Wang2018a}
Z.-F. Wang, J.~Wang, C.-Y. Zeng, Q.-S. Min, Y.~Tian, and M.-Z. Zuo, ``Digital
  audio tampering detection based on enf consistency,'' in \emph{2018
  International Conference on Wavelet Analysis and Pattern Recognition
  (ICWAPR)}.\hskip 1em plus 0.5em minus 0.4em\relax Chengdu: IEEE, Jul. 2018,
  pp. 209--214.

\bibitem{Zeng2023}
C.~Zeng, S.~Feng, D.~Zhu, and Z.~Wang, ``Source acquisition device
  identification from recorded audio based on spatiotemporal representation
  learning with multi-attention mechanisms,'' \emph{Entropy}, vol.~25, no.~4,
  p. 626, Apr. 2023.

\bibitem{Wang2023f}
Z.~Wang, J.~Zhan, G.~Zhang, D.~Ouyang, and H.~Guo, ``An end-to-end transfer
  learning framework of source recording device identification for audio
  sustainable security,'' \emph{Sustainability}, vol.~15, no.~14, p. 11272,
  Jan. 2023.

\bibitem{Chen2023b}
Z.~Chen, M.~Lin, Z.~Wang, Q.~Zheng, and C.~Liu, ``Spatio-temporal
  representation learning enhanced speech emotion recognition with multi-head
  attention mechanisms,'' \emph{Knowledge-Based Systems}, vol. 281, p. 111077,
  Dec. 2023.

\bibitem{Zeng2022a}
C.~Zeng, Y.~Yang, Z.~Wang, S.~Kong, and S.~Feng, ``Audio tampering forensics
  based on representation learning of enf phase sequence,'' \emph{International
  Journal of Digital Crime and Forensics}, vol.~14, no.~1, pp. 1--19, Jan.
  2022.

\bibitem{Wang2022t}
Z.~Wang, Y.~Yang, C.~Zeng, S.~Kong, S.~Feng, and N.~Zhao, ``Shallow and deep
  feature fusion for digital audio tampering detection,'' \emph{EURASIP Journal
  on Advances in Signal Processing}, vol. 2022, no.~69, pp. 1--20, Aug. 2022.

\bibitem{Zeng2021a}
C.~Zeng, D.~Zhu, Z.~Wang, M.~Wu, W.~Xiong, and N.~Zhao, ``Spatial and temporal
  learning representation for end-to-end recording device identification,''
  \emph{EURASIP Journal on Advances in Signal Processing}, vol. 2021, no.~1,
  p.~41, 2021.

\bibitem{Zeng2020}
C.~Zeng, D.~Zhu, Z.~Wang, Z.~Wang, N.~Zhao, and L.~He, ``An end-to-end deep
  source recording device identification system for web media forensics,''
  \emph{International Journal of Web Information Systems}, vol.~16, no.~4, pp.
  413--425, Aug. 2020.

\bibitem{Wang2011a}
Z.-F. Wang, Q.-H. He, X.-Y. Zhang, H.-Y. Luo, and Z.-S. Su, ``Playback attack
  detection based on channel pattern noise,'' \emph{Journal of South China
  University of Technology}, vol.~39, no.~10, pp. 7--12, 2011.

\bibitem{piech2015dkt}
C.~Piech, J.~Bassen, J.~Huang, S.~Ganguli, M.~Sahami, L.~J. Guibas, and
  J.~Sohl-Dickstein, ``Deep knowledge tracing,'' in \emph{NIPS}, vol.~28, 2015.

\bibitem{ghosh2020cak}
A.~Ghosh, N.~Heffernan, and A.~S. Lan, ``Context-aware attentive knowledge
  tracing,'' in \emph{ACM SIGKDD}, 2020, pp. 2330--2339.

\bibitem{nakagawa2019gkt}
H.~Nakagawa, Y.~Iwasawa, and Y.~Matsuo, ``Graph-based knowledge tracing:
  Modeling student proficiency using graph neural network,'' in \emph{2019
  IEEE/WIC/ACM International Conference on Web Intelligence (WI)},
  Thessaloniki, Greece, 2019, pp. 156--163.

\bibitem{kasneci2023chatgpt}
E.~Kasneci \emph{et~al.}, ``Chatgpt for good? on opportunities and challenges
  of large language models for education,'' \emph{Computers and Education:
  Artificial Intelligence}, vol.~4, p. 100179, 2023.

\bibitem{nguyen2022etal}
B.~Nguyen \emph{et~al.}, ``Towards generalized methods for automatic question
  generation in education,'' in \emph{Proceedings of the European Conference on
  Technology Enhanced Learning (ECTEL)}, 2022.

\end{thebibliography}

\end{document}